\documentclass[letterpaper, 10 pt, conference]{ieeeconf}  

\IEEEoverridecommandlockouts                              

\usepackage{cite}
\usepackage{amsmath,amssymb,amsfonts}
\usepackage{graphicx}

\usepackage{subcaption}
\usepackage{textcomp}
\usepackage[table]{xcolor}
\usepackage{xcolor}
\usepackage{multirow}
\usepackage{array}
\usepackage[
linesnumbered,ruled,vlined]{algorithm2e}
\usepackage {algpseudocode}
\usepackage{algorithmicx}
\usepackage{algcompatible}

\newcolumntype{C}{>{\centering\arraybackslash}p{1.8cm}}

\def\BibTeX{{\rm B\kern-.05em{\sc i\kern-.025em b}\kern-.08em
    T\kern-.1667em\lower.7ex\hbox{E}\kern-.125emX}}

\title{ \LARGE \bf A GPT-based Decision Transformer \\for Multi-Vehicle Coordination at Unsignalized Intersections
}

\author{Eunjae Lee, Minhee Kang, Yoojin Choi, Heejin Ahn$^*$%
    \thanks{The Authors are with School of Electrical Engineering, Korea Advanced Institute of Science and Technology,s Daejeon, Republic of Korea.%
        {\tt \small \{eunjae.lee, ministop, choi369j, heejin.ahn\}@kaist.ac.kr}}%
    \thanks{This research was supported by the MSIT(Ministry of Science and ICT), Korea, under the ITRC (Information Technology Research Center) support program (IITP-2024-RS-2023-00259991) supervised by the IITP(Institute for Information \& Communications Technology Planning \& Evaluation), BK21 FOUR(Connected AI Education \& Research Program for Industry and Society Innovation, KAIST EE, No. 4120200113769)}
}

\begin{document}

\maketitle
\thispagestyle{empty}
\pagestyle{empty}

\begin{abstract}
In this paper, we explore the application of the Decision Transformer, a decision-making algorithm based on the Generative Pre-trained Transformer (GPT) architecture, to multi-vehicle coordination at unsignalized intersections. We formulate the coordination problem so as to find the optimal trajectories for multiple vehicles at intersections, modeling it as a sequence prediction task to fully leverage the power of GPTs as a sequence model. Through extensive experiments, we compare our approach to a reservation-based intersection management system. Our results show that the Decision Transformer can outperform the training data in terms of total travel time and can be generalized effectively to various scenarios, including noise-induced velocity variations, continuous interaction environments, and different vehicle numbers and road configurations.

\end{abstract}

\section{Introduction}\label{Sec:Intro}

Intelligent traffic management has been shown to substantially minimize delays, enhance safety, and improve overall mobility. In particular, intelligent management of unsignalized intersections has been the subject of intensive research due to its susceptibility to accidents. 

Research in intelligent intersection management systems is continuously being conducted, with approaches utilizing optimization \cite{dresner2008multiagent, sayin2018information,wuthishuwong2017consensus,tachet2016revisiting,ahn2019abstraction,gholamhosseinian2022comprehensive} and Reinforcement Learning (RL) \cite{liu2022graph, multiagent, intersectrl}. The optimization-based methods use some form of optimization algorithms to find the optimal vehicle trajectories at intersections. However, due to the limitations in computing capabilities, these approaches often resort to heuristics and approximation, such as reservation of intersection resources \cite{dresner2008multiagent, sayin2018information}, finding optimal trajectories with dynamic programming \cite{wuthishuwong2017consensus}, estimations of the arrival time \cite{ tachet2016revisiting}, and the simplification of vehicle dynamics \cite{ahn2019abstraction}.

RL-based methods have been investigated to tackle the issue of computation in intersection management. Once RL models are trained, they are computationally inexpensive and can generate outputs in real-time. For example, in the work of \cite{multiagent} and \cite{intersectrl}, multiple vehicles are coordinated to safely cross the intersection, using a variation of Q-learning and multi-agent deep RL, respectively. 
However, RL-based methods face significant challenges due to their reliance on exploration. This makes it difficult to handle out-of-distribution (OOD) actions and perform multi-task learning, particularly in offline RL settings where exploration is limited, as highlighted by \cite{levine2020offline}.
In addition, traditional RL-based methods struggle with long-term credit assignment, where delayed rewards complicate the learning process.

\setlength{\belowcaptionskip}{-5pt}
\begin{figure}[t] 
\centerline{\includegraphics[width=8cm]{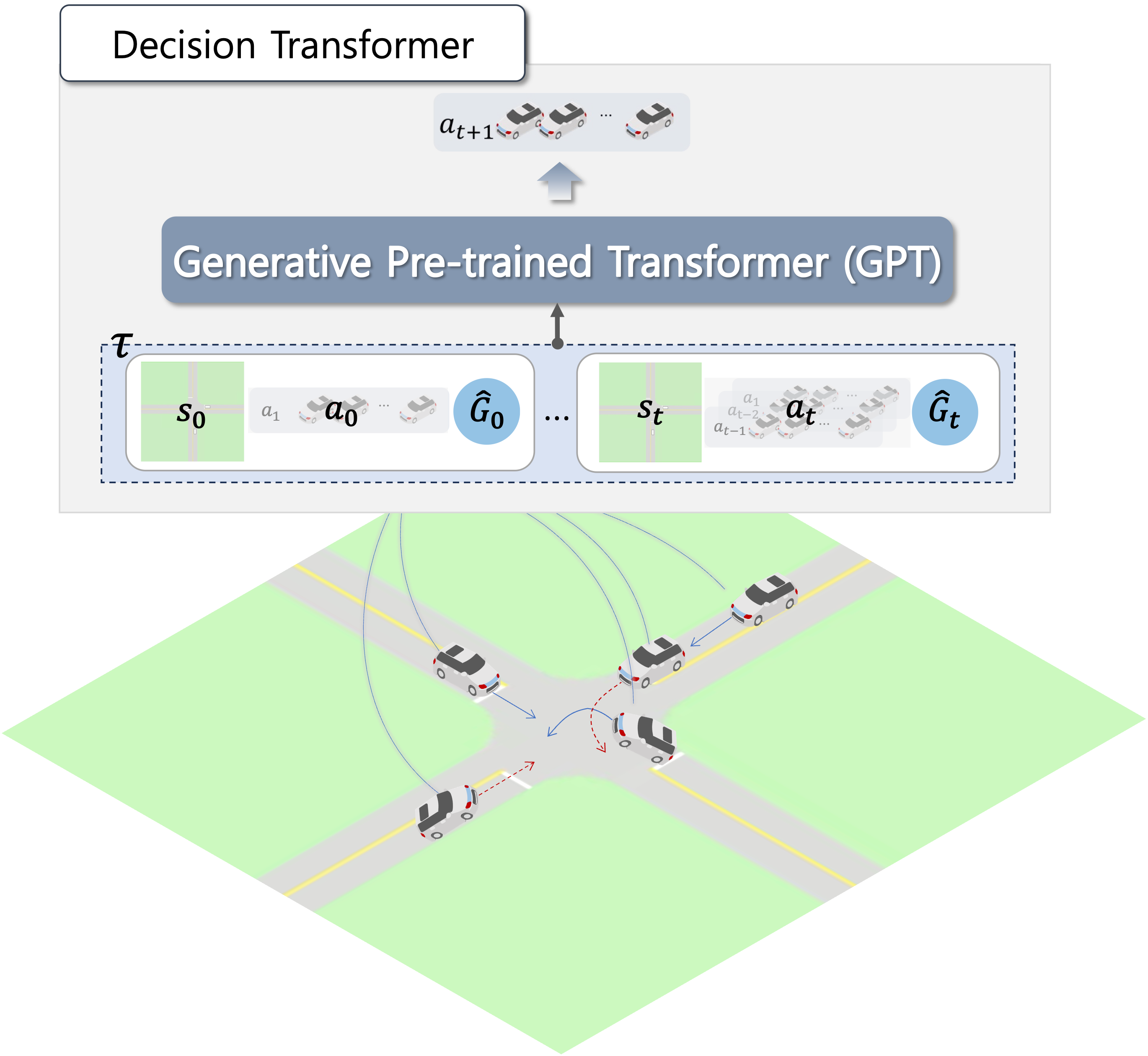}} 
\caption{Multi-vehicle coordination with Decision Transformer}
\label{fig:About this paper2}
\vspace{-10pt}
\end{figure}

In response to these challenges, we turn our attention to the Generative Pre-trained Transformer (GPT) model \cite{radford2019language}, recognizing its potential for decision-making applications (e.g., intersection management) due to its powerful sequential modeling capabilities. These capabilities allow the GPT model to effectively address long-term credit assignment issues. Additionally, the GPT model can generate actions based on historical sequences of states and actions, which enables it to handle out-of-distribution (OOD) actions, support multi-task learning, and derive optimal policies from sub-optimal data.
Although various fields, such as natural language processing, actively utilize the GPT model, it has not yet been carried over to intersection management systems. To the best of our knowledge, the work of~\cite{liu2023mtd} is the most closely related to our work, as they employ the GPT model for single-vehicle control. We extend this approach to address the more complex problem of multi-vehicle coordination.

In this paper, we propose the application of a GPT-embedded decision-making model, the Decision Transformer (DT) \cite{chen2021decision}, to multi-vehicle coordination at unsignalized intersections. To leverage the strengths of the GPT model, we frame the multi-vehicle coordination problem as a sequence prediction task, where the next action is predicted based on the previous sequence of states and actions.
Through extensive experiments, we show that our method can achieve near-optimal performance by learning from sub-optimal data and has the capacity for generalization.

Our contributions are summarized as follows:
\begin{itemize}
\item
We propose the GPT-based Decision Transformer as a solution to the multi-vehicle coordination problem at unsignalized intersections.
\item
We demonstrate that our DT model has the capacity to learn new patterns that can achieve improved performance from sub-optimal data.
\item 
We investigate the generalization ability of our model by highlighting noise resilience and adaptability to various environments. This evidences that the DT can handle scenarios with distribution shifts and multi-tasking.
\end{itemize}

The rest of the paper is organized as follows. We present the problem statement and explain the training and inference processes of the DT in Section \ref{Sec:Method}. In Section \ref{Sec:Experiments}, we experimentally demonstrate the performance of our method in terms of achieving near-optimal solutions from sub-optimal data and noise resilience and adaptability. We conclude the paper in Section \ref{Sec:Con}.

\section{Problem Statement and Solution}\label{Sec:Method}

    In this section, we first define the multi-vehicle coordination problem and present the training and inference algorithms for the DT as a solution to the problem. We also present the data acquisition process.

    \setlength{\belowcaptionskip}{-5pt}
    \begin{figure}[htbp] 
    \vspace{-2pt}
    \centerline{\includegraphics[width=8cm]{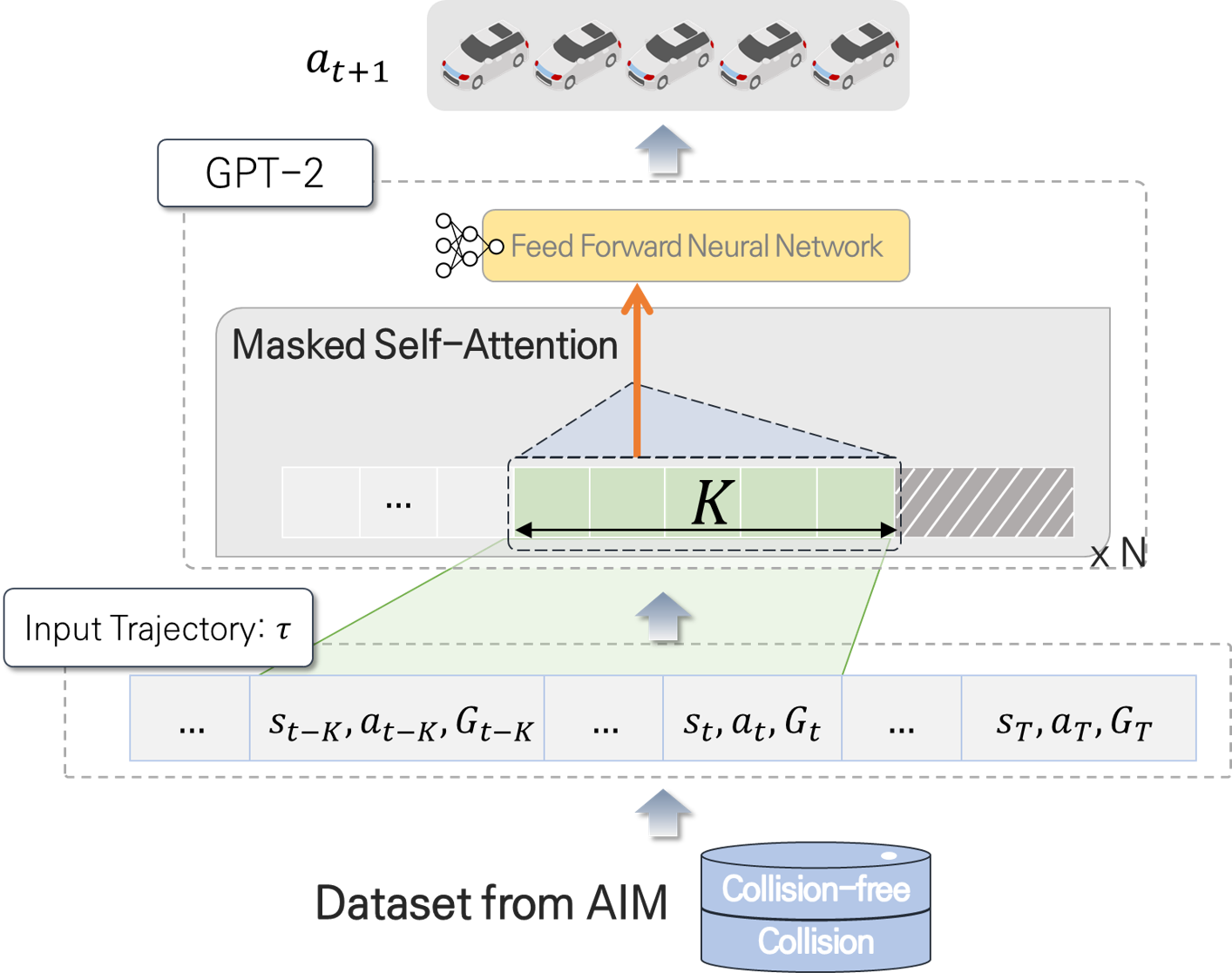}} 
    \caption{Decision Transformer model architecture}
    \label{fig:process}
    \vspace{-2pt}
    \end{figure}
    
  \subsection{Problem Statement}

     We consider the scenario in which an unsignalized intersection environment consists of a single-lane, 4-way intersection with five vehicles as shown in Fig.~\ref{fig:About this paper2}. All five vehicles are randomly generated with the vehicle's entry time, entry location, and destination. We define \textit{one episode} such that it starts when at least one vehicle appears and ends when all five vehicles have passed through the intersection and exited the environment.
     
     We aim to find trajectories of all five vehicles that minimize the time they pass through the intersection while minimizing collisions between them.

    \subsection{Decision Transformer}
    
    The Decision Transformer (DT)~\cite{chen2021decision} is a novel model that combines the GPT with offline RL to predict optimal actions based on historical sequences of states, actions, and returns. 
    Unlike traditional offline RL, which infers actions from future reward expectations, the DT \textit{generates} the next action by leveraging past sequences. This approach enables the model to handle wide distributions of behaviors, which is typically challenging for conventional RL methods.

    The GPT architecture traditionally generates sequential outputs, such as texts, by conditioning on a prompt. In the context of our DT, the prompt corresponds to a trajectory as shown in Fig.~\ref{fig:process}, where a trajectory consists of states, actions, and target returns, also known as Return-To-Gos (RTGs). 
    Let $T$ denote the length of an episode, and let $s = (s_{1}, s_{2}, \cdots, s_{T})$, $a = (a_{1}, a_{2}, \cdots, a_{T})$, and $G = (G_{1}, G_{2}, \cdots, G_{T})$ denote the sequence of states, actions and RTGs, respectively. Then, a trajectory $\tau$ is defined as 
    \begin{equation} \label{eq:input}
        \tau := (s_1,a_1,G_1, \dots,s_T,a_T,G_T).
    \end{equation}
    In the following, we explain how we define the states, actions, and RTGs in the vehicle coordination problem.    
        \paragraph{States}
        The state consists of the position, velocity, heading angle, and destination of all vehicles. We assume that the environment is fully observed, i.e., the observation space is equivalent to the state space. We denote the state of vehicle $i$ at time $t$ as         \begin{equation}\label{eq:state}
            s_{i,t} := (x_{i,t},y_{i,t}, v_{i,t}, \psi_{i,t}, x_{i,des}, y_{i,des}),
        \end{equation}
        where $(x, y)$ is the ₩position in a two-dimensional Cartesian coordinate system, $v$ and $\psi$ are the velocity and heading angle, respectively, and $(x_{i,des}, y_{i,des})$ denotes the vehicle's destination. 
        For all five vehicles, the state is constructed as $s_t =(s_{1,t}, s_{2,t}, \cdots , s_{5,t})$.
        
        \paragraph{Actions}
        We define the action $a_{i,t}\in[-1.5, 1.5]$ for vehicle $i$ at time $t$ as a continuous acceleration that vehicle $i$ applies for one time step. We predetermine the paths of all vehicles to reach their destination $(x_{i,des}, y_{i,des})$ and consider the longitudinal acceleration as the action.

        \paragraph{Return-To-Gos}
        The RTG, which represents the cumulative sum of future rewards, is a key component in the DT. By using the RTG as input, the DT can learn behavior patterns that maximize rewards, and generate the optimal actions based on the RTG.
        
        In the vehicle coordination problem, we aim to make vehicles pass through the intersection as fast as possible without collisions. Thus, the reward should increase as vehicles' speeds increase and be penalized when vehicles collide. we define the reward for vehicle $i$ at time $t$ as 
        \begin{equation}
            r_{i,t}:= c_1\cdot \frac{v_{i,t}-v_{min}}{v_{max}-v_{min}} - c_2\cdot collsion. \label{eq:reward}
        \end{equation}
         The parameters $c_1$ and $c_2$ represent the weights of the speed and safety factors, respectively. Here, $v_{max}$ denotes the speed limit of the road, while $v_{min}$ denotes the minimum speed of the vehicle, which in our case is 0 as vehicles should not be moving in reverse. If there is a collision in the episode, $collision$ is set to 1, and 0 otherwise. For all five vehicles, the reward at time step $t$ is the sum of the individual rewards, that is, $r_t := r_{1,t} + r_{2,t}+ \ldots + r_{5,t}$.
        
        The RTG $G_t$ is defined as the sum of future rewards from a given time step $t$ to the end of the episode $T$, that is,
        \begin{equation}
         {G}_t = \sum_{k=t}^T r_k. \label{eq:rtg}
        \end{equation}
During training, the RTG is computed precisely from offline datasets, while during inference, it is estimated. We will explain the estimation of the RTG during the discussion of Algorithm~\ref{alg:pseudocode_eval}.

    We train the DT using trajectories as shown in Fig.~\ref{fig:process}, and Algorithm~\ref{alg:pseudocode_training} details the training process. 
    The DT model consists of a positional encoding layer, and $N$ layers of a masked self-attention and a feedforward neural network like GPT-2~\cite{radford2019language}. The positional encoding layer is used to preserve the sequential information. In our case, we use an positional encoder to preserve the order of trajectories, and inform the model that $s_t, a_t$, and $G_t$ are at the same time step $t$. The masked self-attention layer helps the model learn the relationship of input trajectories. Through these components, the model is trained to generate the next action autoregressively and learn meaningful patterns from historical trajectories from the masked future trajectory. Also, the DT uses the mean-square error (MSE) between the generated action $a_{t+1,\textrm{pred}}$ and the actual action $a_{t+1}$ during training to update the model's parameters.

    Algorithm~\ref{alg:pseudocode_eval} details the evaluation phase, which is depicted in Fig.~\ref{fig:About this paper2}. The DT introduces an initial estimate RTG $\hat{G}_0$ as an input to guide decision-making. The quality of the initial estimate RTG is crucial, and its value is typically chosen as an average of the sum of all rewards in the training data or determined experimentally.
    After the DT generates the action, the RTG is updated through $\hat{G}_{t} = \hat{G}_{t-1} - r_{t}$ where $r_t$ is an observed reward. The RTG acts as a goal for the DT, ensuring that the generated actions lead to the maximization of rewards over the remaining time steps. 
    This process repeats until a termination condition is satisfied, where the termination condition is when a collision occurs or when all five vehicles have passed through the intersection.

    \begin{algorithm}
    \caption{Decision Transformer Training} \label{alg:pseudocode_training}
    
    \textbf{Input:} Trajectories from AIM Dataset $D$, Context Length $K$, Iterations $iter$, Steps per Iteration $step$
    
    \textbf{Output:} Trained Model \textsc{DecisionTransformer}     
    
    \For{$iteration = 1$ to $iter$} {
        \For{$step~per~iteration = 1$ to $step$}{
             \textbf{Data Sampling: }$\tau = (s_{t-K}, a_{t-K}, G_{t-K}, \cdots, s_{t}, a_{t}, G_{t})$ from dataset $D$  
            
            \textbf{Positional encoding: }$\tau_e$ = \textsc{Positional encoder($\tau$)}
            
            \textbf{Training: }$a_{t+1,\text{pred}} \gets \textsc{DecisionTransformer}(\tau_e)$

            \textbf{Update model parameters: } Use gradient descent with MSE loss of $a_{t+1} - a_{t+1,pred}$ where $a_{t+1}$ is the actual action in dataset $D$
            
        }
    }
    \end{algorithm}

\begin{algorithm}
\caption{Decision Transformer Evaluation}    
\label{alg:pseudocode_eval}

 \textbf{Input: } Trained Model \textsc{DecisionTransformer}, Initial estimate RTG $\hat{G}_0$, Initial state $s_0$, Max episode Length $T_{\max}$
 
 \textbf{Output: }Evaluated trajectory $\tau$ 
 
 Initialize trajectory $\tau = (s_0, \emptyset, \hat{G}_0)$

\For{$t = 1$ to $T_{max}$}{
     \textbf{Action generation: } $a_{t} \gets   \textsc{DecisionTransformer}(\tau)$
     
     Execute $a_{t}$ and observe new state $s_{t}$ and reward $r_{t}$
     
     \textbf{Update RTG: } $\hat{G}_{t} \gets \hat{G}_{t-1} - r_{t}$   
     
    \If{\text{Termination condition is met}} {
         \textbf{Break loop} 
    }
     \textbf{Update trajectory: } $\tau \gets (\tau, s_t, a_t, \hat{G}_t)$}

\end{algorithm}

\subsection{Data Acquisition} 

    In training our DT model, we use two sets of data: vehicles crossing the intersection collision-free at high efficiency and vehicles crossing the intersection as fast as possible with no regard for collisions. Currently, no such real-world datasets for both conditions are available, and thus, we acquire data through simulation implementations of effective intersection management methods. The simulations were run in MATLAB~\cite{mathworks2022matlab}, with their respective scenarios implemented for both data types.

    \subsubsection{Collision-free dataset}

    To learn the correct behavior patterns and decision rules, we collect collision-free data. To this end, we use the reservation-based AIM \cite{dresner2008multiagent}, an effective and widely researched method for intersection management. This method operates by directly managing a grid of space-time in the interior of the intersection. A vehicle approaching the intersection attempts to reserve slots in space-time that overlap with its predicted trajectory. A reservation is accepted if it does not conflict with existing reservations, in which case the vehicle follows its predicted trajectory. However, a reservation is rejected if it conflicts with existing reservations, and the vehicle should attempt to make a reservation again at a later time. The reservation-based method has proven to be effective in significantly reducing travel time delays of over 99\% for vehicles passing through an intersection \cite{dresner2008multiagent}.

    \subsubsection{Collision dataset}

    To prevent collision situations, the model should learn some collision patterns. 
To collect such data, we do not implement any form of vehicle coordination, allowing vehicles to travel at maximum velocity.

\section{Experiments}\label{Sec:Experiments}

In this section, we consider the number of training datasets and training data composition to obtain the best-trained DT model. Then, we compare the DT model with the reservation-based AIM to demonstrate that our model can learn new patterns that achieve improved performance. Lastly, we explore the generalization properties of our method towards noise resilience and adaptability.

Throughout this section, we use three evaluation metrics: episode return, episode length, and collision rate. The episode return refers to the summation of all rewards, and the episode length refers to the end time of the episode. 
The collision rate is the number of episodes with collisions divided by number of total episodes.

\subsection{Model Training}

    To find the best-trained model, we set the hyper-parameters of iteration $iter$, step per iteration $step$, context length $K$, the number of attention heads $H$, the number $N$ of transformer layers (multi-head attention and feedforward neural network), and the embedding dimension $dim$ of the input data. 
    As the depth of the multi-head attention increases, the model can better focus on different parts of the sequence, enhancing pattern recognition at the cost of increasing complexity. A large number of transformer layers deepen contextual understanding, helping with complex tasks at the cost of increasing training difficulty.
    The embedding dimension determines the richness of the feature representation, balancing between capturing enough detail and avoiding overfitting or excessive computation. Therefore, we set the hyper-parameters as $iter=50,step=10,000, K=30,H=8,N=12$, and $dim=128$.

    \paragraph{Mixed with Collision Data}
    To find a suitable data composition, we compare the model trained only with the collision-free dataset and the other models trained with mixed datasets. The collision-free dataset contains 512,000 collision-free trajectories, which is composed of 500 trajectories from each entry combination for 1,024 total combinations, and the three mixed datasets consist of the collision-free dataset (512,000 data) combined with extra collision data numbers of 5\%, 10\%, and 20\%  of the number of collision-free data, respectively.

    To validate the performances of the models, we randomly select 100 test cases not included in the training dataset, with the collision episodes excluded. As shown in Table~\ref{table:compare collision data}, all the models show similar average lengths, but the model trained by the 10\% mixed dataset exhibits the lowest collision rate. Since our goal is to minimize passing time and collisions, we utilize the model with the 10\% mixed dataset throughout the experiments.

    \setlength{\belowcaptionskip}{5pt}
    \begin{table}[htbp]
    \caption{Model performance with different proportions of collision data. `With x\%' indicating the mixed dataset containing x\% collision data.}
    \vspace{-2pt}
    \renewcommand{\arraystretch}{1.3} 
    \begin{center}
    \begin{tabular}{|c|c|c|c|}
    \hline
    \multirow{2}{*}{Datasets}  & \multirow{2}{*}{\textbf{Avg Return $\uparrow$}}& \multirow{2}{*}{\textbf{Avg Length $\downarrow$}} & \multirow{2}{*}{\textbf{Collision Rate $\downarrow$}} \\ 
    & & & \\
    \hline
    \textbf{Collision-free} & {1259.64} &	{18.78} & {0.14}  \\
    \hline
    \textbf{With 5\%} & {1258.99} &	{\textbf{18.57}} & {0.12}  \\
    \hline
    \textbf{With 10\%} & {1259.36} & {18.67} &	\textbf{{0.08}} \\
    \hline
    \textbf{With 20\%} & {\textbf{1259.84}} & {18.70} & {0.12} \\
    \hline
    \hline
    \textbf{AIM} & {1259.25} &	{18.20} & {0.0}  \\
    \hline
    \end{tabular}
    \label{table:compare collision data}
    \end{center}
    \vspace{-10pt}
    \end{table}

\subsection{Comparison with Optimal Solution}
\begin{table*}[htbp]
    \centering
    \setlength{\belowcaptionskip}{5pt}
    \caption{Results of the near-optimal approach: Best indicates scenarios where DT outperforms AIM system in terms of Length with approaching optimal solution, and Worst indicates scenarios where AIM outperforms DT.}
    \begin{tabular}{|c|c|c|c|c|c|c|c|}
        \hline
        &\multirow{2}{*}{\textbf{Scenario Index}} & \multicolumn{2}{c|}{\textbf{Decision Transformer}} & \multicolumn{2}{c|}{\textbf{AIM}}  & \textbf{Optimal Solution} \\
        \cline{3-7}
        & & \textbf{Return} & \textbf{Length (s)} & \textbf{Return} & \textbf{Length (s)}  & \textbf{Length (s)}\\
        \hline
        \hline
        \multirow{7}{*}{\textbf{Best}} &\cellcolor{lightgray}1 & \cellcolor{lightgray}1253.01	&\cellcolor{lightgray}\textbf{17.4}	&\cellcolor{lightgray}1254.50&	\cellcolor{lightgray}\textbf{21.4} &\cellcolor{lightgray}\textbf{16.8}\\
            \cline{2-7}
            &2 & 1276.62&	17.4	&1276.04	&19.8& {16.6}  \\
            \cline{2-7}
            &3 & 1241.55	&15.6	&1243.74	&17.5& {15.6} \\
            \cline{2-7}
            &\multicolumn{6}{c|}{ \vdots }\\
            \cline{2-7}
            &14 & 1241.40&	16.7&	1243.7&	16.8 & 15.7 \\
            \cline{2-7}
            &15 &1239.88& 17.5	& 1240.33 &  17.5 & 15.2 \\
        \hline
        \multicolumn{2}{|c|}{\textbf{Best Scenario Average (std)}} & 1258.9 (16.85)& 17.64 (0.93) & 1259.75 (16.23) & 18.42 (1.32)& - \\
        \hline
        \hline
        \multirow{6}{*}{\textbf{Worst}} 
            &16 & 1241.74	& 16.8 & 1242.43	& 16.7 & 15.9\\
            \cline{2-7}
            &17 & 1267.00 &	 20.2 &	1267.44	& 20.0 & 15.3 \\
            \cline{2-7}
            &\multicolumn{6}{c|}{ \vdots }\\ 
            \cline{2-7}
            &\cellcolor{lightgray}90 & \cellcolor{lightgray}1236.13 &\cellcolor{lightgray}23.3 &\cellcolor{lightgray}1258.97&\cellcolor{lightgray}18.7& \cellcolor{lightgray}15.6 \\
            \cline{2-7}
            &91 & 1236.13 &	20.8&	1235.40&	16.1 & 15.8 \\
        \hline
        \multicolumn{2}{|c|}{\textbf{Worst Scenario Average (std)}} & 1256.65 (12.64)& 18.93(1.70) & 1255.59 (12.22)& 17.61 (1.26)& - \\
        \hline
        \hline
        \multicolumn{2}{|c|}{\textbf{Total Scenario Average (std)}} & 1257.02 (13.34)& 18.72 (1.66) & 1256.27 (12.96) & 17.75 (13.10)& - \\
        \hline
    \end{tabular}
    \label{tab:optimal example}
\end{table*}

\begin{figure*}[htbp]
\hspace{0.7cm} 
   \begin{subfigure}{0.40\linewidth}
       \centering
       \includegraphics[width=\textwidth]{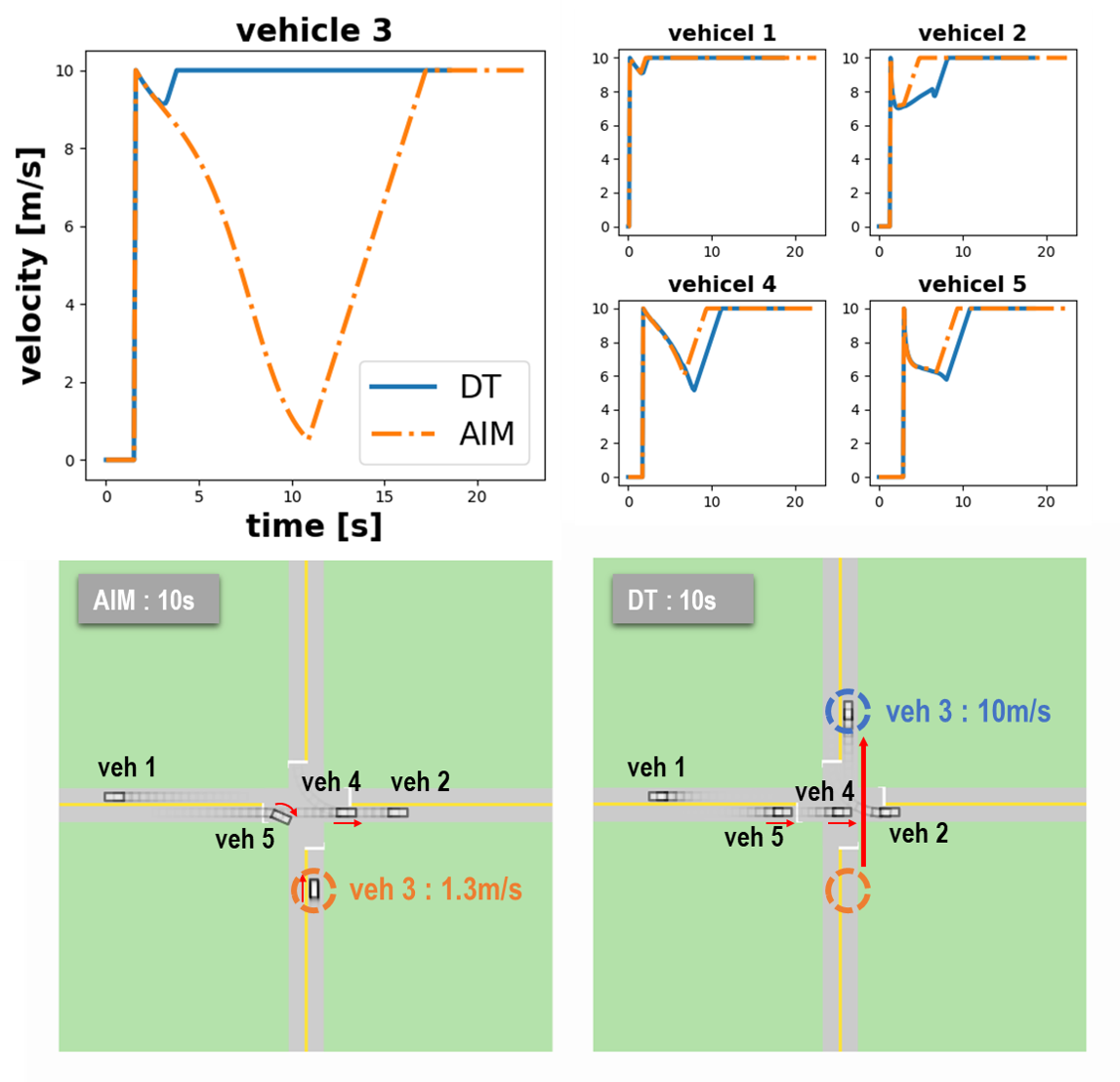}
       \caption{Scenario \#1 where our DT model generates the near-optimal solution.}
       \label{fig:image1}
   \end{subfigure}
   \hfill
   \begin{subfigure}{0.40\linewidth}
       \centering
       \includegraphics[width=\textwidth]{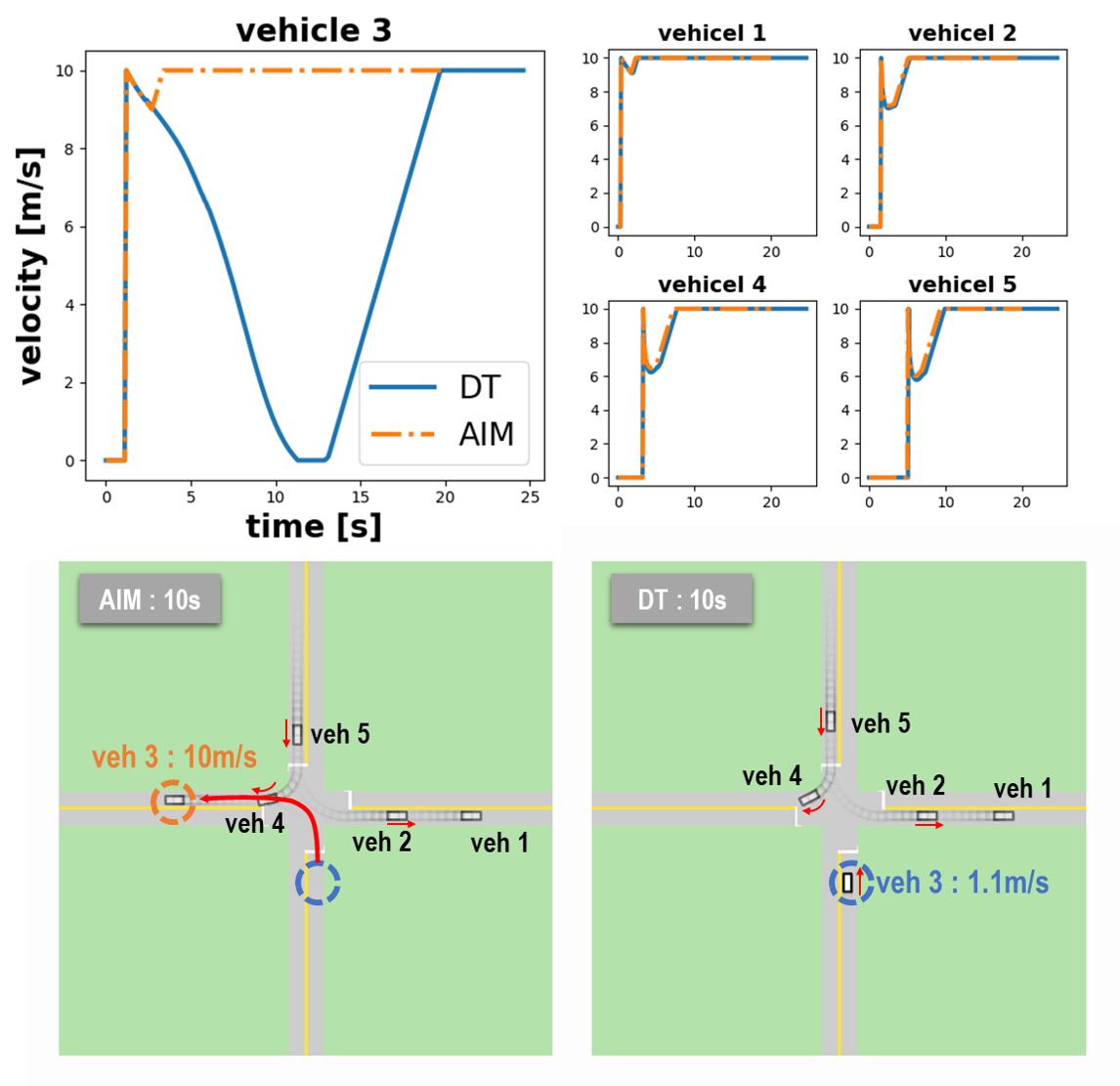}
       \caption{Scenario \#90 where our DT model generates a new solution but yields a worse performance.}
       \label{fig:image3}
   \end{subfigure}
\hspace{0.7cm} 
   
   \hfill
    \caption{Visualization of the driving trajectories of our model and AIM with vehicle speed trend at 2 cases }
    \label{fig:optimal}
    \vspace{-7pt}
\end{figure*}


We compare our DT model with the reservation system and the optimal solution. In Section \ref{Sec:Experiments}.B, we use 100 cases from the training dataset to assess the DT's ability to generate new solutions when it is trained with the same training dataset. We compute the optimal solution using the optimal scheduling approach \cite{gholamhosseinian2022comprehensive}. In particular, we consider all possible orders of five vehicles passing through the intersection and find the solution with the shortest total delay. The optimal scheduling approach has the same conditions as \textbf{}the reservation-based system, but safety distance and buffer are minimized to calculate the maximum performance.

Table~\ref{tab:optimal example} presents the episode return and length of 91 scenarios where 9 scenarios are excluded due to collisions. In 15 scenarios, the DT shows shorter episode lengths than the reservation-based AIM.
Specifically, we highlight two scenarios, scenario \#1, which shows the largest performance improvement, and scenario \#90, which shows the largest performance degradation. In scenario \#1, our DT outperforms the reservation-based AIM system with a length reduction of 4s and the episode length is close to the optimal solution, showing a difference of 0.6s. In scenario \#90, the episode length of our DT is longer by 4.6s than the reservation-based AIM system.

We visualize both scenarios in Fig.~\ref{fig:optimal} for detailed inspection. Fig.~\ref{fig:optimal}(a) depicts scenario \#1 where the DT enables vehicle 3 to pass through the intersection before the other vehicles (vehicles 4 and 5), whereas the AIM system makes it yield. 
Fig.~\ref{fig:optimal}(b) depicts scenario \#90 where the AIM system lets vehicle 3 accelerate and pass through the intersection, whereas the DT makes it yield to vehicles 4 and 5.
It is noteworthy that our DT generates priority rules and driving patterns that are different from those in the training dataset. Although new patterns may lead to sub-optimal performance, it is promising that they can also lead to near-optimal solutions by learning from sub-optimal data.

\subsection{Generalization}
 
    To demonstrate the generalization of the DT, we test it in various situations, including velocity noise, continuous traffic flow, and variation in vehicle numbers and road configurations. The noise introduces a domain shift, and variations create tasks not encountered during training.
    
    All the results in this section are obtained by evaluating our DT model on 100 new test cases that are not in the training dataset.

    \paragraph{Noise Resilience}  
    We introduce an evaluation environment by injecting random noise, ranging up to 2\%, into the velocity according to \cite{UNECE2023}. Due to the addition of noise, vehicles should respond with different patterns of behavior, unlike those of situations in which the vehicle's behavior had previously learned. 

\begin{figure}[htbp]
    \begin{subfigure}{1\linewidth}
       \centering
       \includegraphics[width=\textwidth]{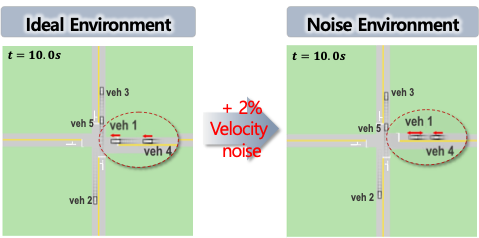}
       \caption{Visualization of the model evaluation process at 10s.}
   \end{subfigure}
    \hfill 
    
   \begin{subfigure}{1\linewidth}
       \centering
       \includegraphics[width=\textwidth]{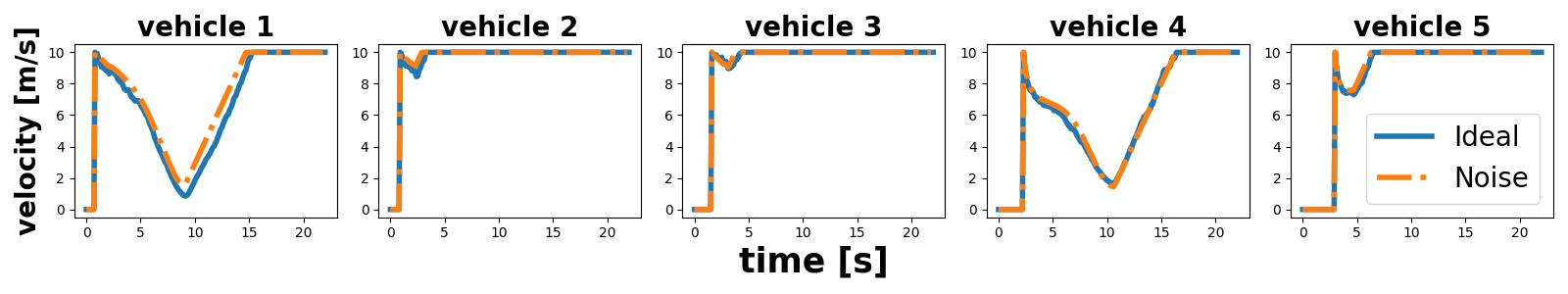}
       \caption{Velocity of 5 vehicles comparison.}
   \end{subfigure}
   \hfill 
    \caption{Environments with noise-free and noise.}
    
    \label{fig:noise}
    \end{figure}

Table~\ref{table:Noise} shows the results in all three metrics. They all exhibit a gentle decline, with an average return decline of 0.08\%, an average length increase of 1.5\%, and a collision rate increase of 6\% in the noisy environment. 
    As depicted in Fig.~\ref{fig:noise}(a), in the ideal environment, vehicles 1 and 4 maintain a safe distance, avoiding collisions. In contrast, in a noisy environment, the distance between the two vehicles is remarkably close, potentially resulting in a collision. While the other vehicles have similar speeds in both ideal and noise environments, vehicle 1 exhibits minor delays in speed, as shown in Fig.~\ref{fig:noise}(b). 
{Despite the noise, the model learns to adjust the safety distances while maximizing the rewards.}
    
    This suggests that the DT has the capability to control multiple vehicles in an environment that is different from the training environment, that is, in a distribution-shifted environment.
    
    \begin{table}[htbp]
    \setlength{\belowcaptionskip}{-5pt}
    \caption{Model performance in environments with and without noise.}
    \vspace{-5pt}
    \renewcommand{\arraystretch}{1.5} 
    \begin{center}
    \begin{tabular}{|c|c|c|c|}
    \hline
    \textbf{Noise} & \textbf{Avg Return}& \textbf{Avg Length} & \textbf{Collision Rate} \\ 
    \cline{2-4}
    \hline
    \textbf{w/o} & {1259.36} & {18.67} &	{{0.08}} \\
    \hline
    \textbf{w}  & 1258.04 & 18.49 & 0.14 \\
    \hline
    \end{tabular}
    \label{table:Noise}
    \end{center}
    \vspace{-10pt}
    \end{table}

    \paragraph{Continuous traffic flow} 
        
    To evaluate whether the DT model is able to handle a continuous vehicle flow, we build an evaluation environment in which five vehicles constantly enter and interact at the intersection for 300 seconds. Table~\ref{table:continue} shows the average return per vehicle and collision rates between the single interaction environment (the original environment) with a continuous interaction environment. The DT model shows similar performance for both environments, suggesting that the DT model trained from learning single interaction environments can adapt to the continuous environment.

    \begin{table}[htbp]
    \caption{Model performance in continuous traffic flow.}
    \vspace{-5pt}
    \renewcommand{\arraystretch}{1.5} 
    \begin{center}
    \begin{tabular}{|c|c|c|}
    \hline
    \textbf{Interaction} & \textbf{Return per vehicle } & \textbf{Collision Rate} \\ \hline
Single                                & 251.8           & 0.08                    \\ \hline
Continuous                            & 251.7           & 0.10                   \\ \hline
    \end{tabular}
    \label{table:continue}
    \end{center}
    \vspace{-10pt}
    \end{table}

    
    \paragraph{Transferability to various environments}  
    To identify the potential to expand into different environments, we build two evaluation environments, one involving three vehicles and the other with a 3-way intersection. 
    
    In the environment with three vehicles, the number of vehicles is reduced while maintaining the same 4-way intersection environment. As shown in Table.~\ref{table:Variations}, 
    in the three-vehicle environment, the DT exhibits a lower collision rate and allows for faster passage through intersections than in the five-vehicle environment.
    Our model demonstrates the ability to handle multi-tasking, successfully coordinating scenarios of both five and three vehicles. Also, the results imply that once the model is trained in an environment with $n$ vehicles, it can effectively adapt to environments with fewer vehicles.
    
    In the 3-way intersection scenario, the road configuration is changed to a 3-way intersection while keeping the number of vehicles at five. Table~\ref{table:Variations} indicates a slight difference in average return and a consistent collision rate compared to the 4-way intersection. Despite the increased frequency of interactions in the 3-way setup with the same number of vehicles, the results demonstrate that our DT model can adapt effectively to different intersection configurations.

  In summary, the DT model exhibits generalization capabilities in various environments, demonstrating that it can handle OOD scenarios that are unseen during training and support multi-tasking.


\begin{table}[htbp]
\caption{Model performance in variations of vehicle numbers and road configuration. The value of average return means the average return per vehicle, and the values in parentheses indicate the total return on all the vehicles.}
\vspace{-5pt}
\renewcommand{\arraystretch}{1.5} 
\begin{center}
\begin{tabular}{|c|c|c|}
\hline
\textbf{Vehicle Numbers}    & \textbf{Avg Return} & \textbf{Collision Rate} \\ \hline
5 vehicles                  & 251.8 (1259.05)     & 0.08                    \\ \hline
3 vehicles                  & 252.3 (756.91)      & 0.04                    \\ \hline \hline
\textbf{Road Configuration} & \textbf{Avg Return} & \textbf{Collision Rate} \\ \hline
4-way Intersection          & 251.8 (1259.05)                   & 0.08                       \\ \hline
3-way Intersection          & 251.3 (1256.28)                   & 0.08                   \\ \hline
\end{tabular}
\label{table:Variations}
\end{center}
\end{table}

\section{CONCLUSION}\label{Sec:Con}
 In this paper, we have proposed the application of the GPT-based decision-making method, the Decision Transformer, to intersection management. We have demonstrated the ability of the DT to derive new patterns that can achieve near-optimal performances.
 Also, we have shown that the DT can adapt to new environments, including untrained environments with noise, continuous vehicle flow environments, and variations in vehicle numbers and road configurations. 
 Based on the promising results that we present in this paper, we plan to 
 design a safety filter that monitors the action generated by the DT and corrects it if it leads to collisions. 
 Also, by leveraging the generalization capabilities of the DT, we will extend it to handle intersection management when human-driven vehicles co-exist.


\vspace{12pt}

\bibliographystyle{IEEEtran} 
\bibliography{reference} 

\end{document}